%% file: 0_main.tex
\documentclass{article}
\usepackage{ijcai24}
\usepackage{times}
\usepackage{soul}
\usepackage{url}
\usepackage[hidelinks, colorlinks, linkcolor=Black, anchorcolor=BrickRed, citecolor=Black, urlcolor=Black]{hyperref}
\usepackage[utf8]{inputenc}
\usepackage[small]{caption}
\usepackage{graphicx}
\usepackage{amsmath}
\usepackage{amsthm}
\usepackage{booktabs}
\usepackage{algorithm}
\usepackage{algorithmic}
\usepackage[switch]{lineno}
\usepackage{multirow}
\usepackage{latexsym}

\input{framework_setting}

\title{A Survey on Game Playing Agents and Large Models: \\Methods, Applications, and Challenges}
\author{
Xinrun Xu$^{2,3}$\and
Yuxin Wang$^4$\and
Chaoyi Xu$^5$\and\\
Ziluo Ding$^1$\and
Jiechuan Jiang$^6$\and 
Zhiming Ding$^2$\and
Börje F. Karlsson$^1$\footnote{Corresponding Author}\\
\affiliations
$^1$Beijing Academy of Artificial Intelligence (BAAI)
$^2$Institute of Software, Chinese Academy of Science\\
$^3$University of Chinese Academy of Science
$^4$Dartmouth College\\
$^5$Beijing University of Posts and Telecommunications
$^6$Peking University
\emails
xuxinrun20@mails.ucas.ac.cn,
borje@baai.ac.cn
}

\begin{document}

\maketitle

\begin{abstract}
The swift evolution of Large-scale Models (LMs), either language-focused or multi-modal, has garnered extensive attention in both academy and industry. But despite the surge in interest in this rapidly evolving area, there are scarce systematic reviews on their capabilities and potential in distinct impactful scenarios. This paper endeavours to help bridge this gap, offering a thorough examination of the current landscape of LM usage in regards to complex game playing scenarios and the challenges still open. Here, we seek to systematically review the existing architectures of LM-based Agents (LMAs) for games and summarize their commonalities, challenges, and any other insights. Furthermore, we present our perspective on promising future research avenues for the advancement of LMs in games.
We hope to assist researchers in gaining a clear understanding of the field and to generate more interest in this highly impactful research direction.
A corresponding resource, continuously updated, can be found in our GitHub repository\footnote{ \href{https://github.com/BAAI-Agents/GPA-LM}{https://github.com/BAAI-Agents/GPA-LM}}.   
\end{abstract}

\section{Introduction}\label{sec:intro}

\begin{figure*}
    \centering
    \includegraphics[width=\linewidth]{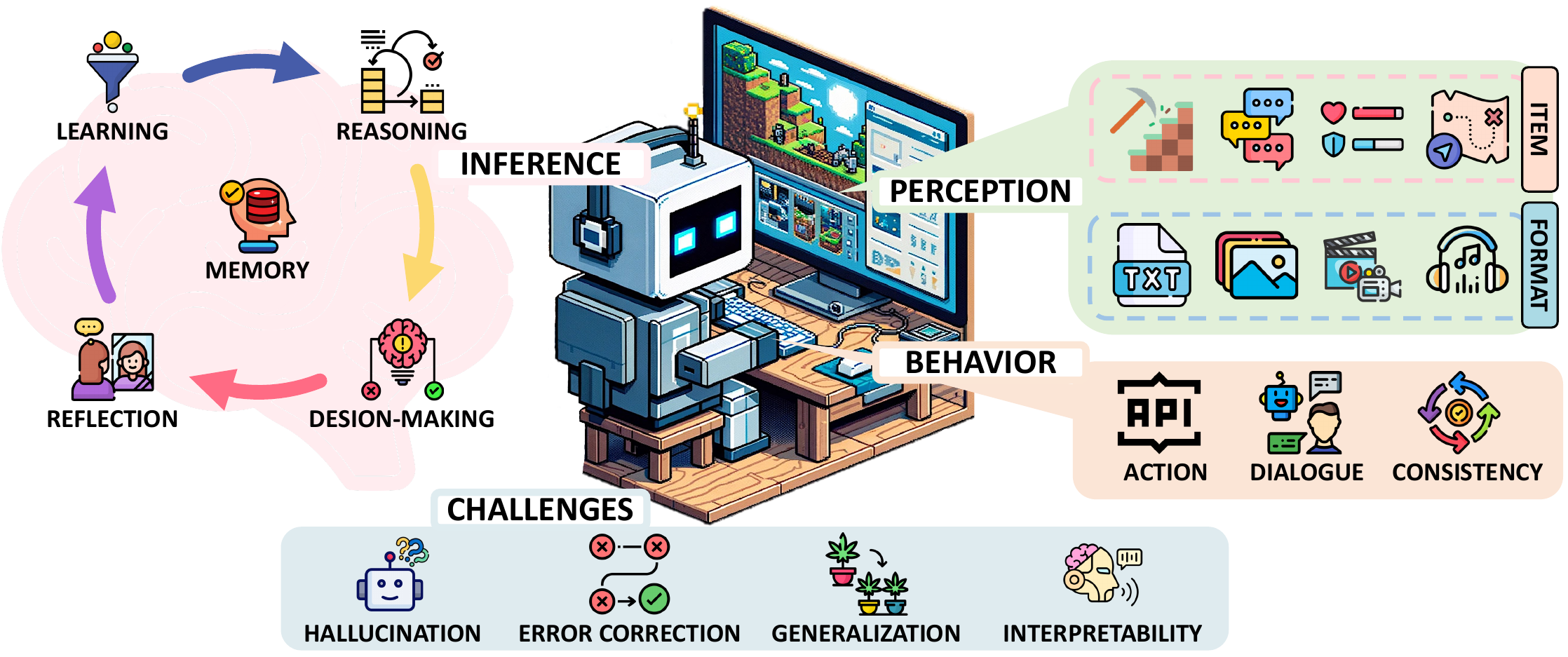}
    \caption{Target areas and Challenges leveraging Large Model capabilities in Game Playing Agents.} 
    \label{fig:diagram}
\end{figure*}

\input{tree}

The development of large-scale models (LMs), including both language and multi-modal models, has been a significant advancement in the fields of natural language processing and computer vision. Recent advancements in LMs have led to notable achievements across various applications, including text generation \cite{zhu2023calypso}, image understanding \cite{zheng2023steve}, and robotics \cite{driess2023palm}.  
This progress has led researchers to explore the application of LMs "as" agents to perform complex tasks, where LM-based Agents (LMAs) in many cases have exhibited interesting generalization capabilities compared to their traditionally trained counterparts \cite{anonymous2024true,wang2023describe}. 
The demonstrated capabilities of LMs have led to considerable interest in their application to game playing. This interest is particularly evident in popular games such as Minecraft \cite{lifshitz2023steve}, 
where LMs' potential to handle complex, dynamic environments is being actively explored.

In the context of pursuing Artificial General Intelligence (AGI) research, digital games are recognized as significant for their provision of complex challenges that necessitate advanced reasoning and cognitive abilities, serving as an ideal benchmark for assessing agent and system capabilities \cite{wu2023smartplay,bellemare2013arcade}. The process of data acquisition in gaming contexts offers advantages in terms of cost-effectiveness, controllability, safety, and diversity in comparison to real-world experiments, while preserving significant challenges. 
Although attempting to analyze or formalize game AI agents and their components is hardly a recent phenomena even outside academia~\cite{IGDA2003,IGDA2004,IGDA2005}, investigating the performance of LMAs within complex gaming environments is crucial for delineating their current limitations and assessing the progress towards autonomy, generalizability, informing the design of new architectures, and moving closer to a potential AGI.  
Furthermore, this survey represents an inaugural comprehensive examination of Game Playing Agents and Large Models, aiming to catalyze subsequent research in this area, by providing an overview of recent endeavors that integrate LMAs with gaming playing applications (e.g., as main player, assisting human players, controlling NPCs), contextualizing and contrasting them, and identifying remaining open challenges.

The human brain functions as a sophisticated information processing system that first transforms sensory information into perceptual representations, then uses these to construct knowledge about the world and make decisions, and finally implements decisions through action \cite{cisek2010neural}. As this abstract sequence mirrors the typical iteration cycle observed in game-playing agents, which consists of perception (\S\ref{sec:perception}), inference (\S\ref{sec:brain}), and action (\S\ref{sec:action}), we follow a similar organization in this survey. \autoref{fig:diagram} illustrates the core survey structure, covering the essence of how sensory information is converted into actions and how LMs can play a role in each step. 

\noindent \textbf{Perception} involves transforming raw observation information during game play into actionable insights, supporting subsequent interactions. Initial studies focused on understanding semantic information through text~\cite{wang2023voyager,xu2023exploring}, while more recent work focuses on integrating visual information (e.g.,~\cite{zheng2023steve}). 

\noindent \textbf{Inference} encompasses key abilities of game agents including memorization, learning, reasoning, reflection, and decision-making; usually built upon a comprehensive cognitive framework. 
The general framework is adaptable to various application contexts, meaning \emph{not all individual components are necessary in every scenario.}
Specifically, memorization is about efficient storage and retrieval of learnt knowledge to enhance common sense and game-specific insights~\cite{anonymous2023ghost,hong2023metagpt}. 
Learning usually concerns skill acquisition and strategic adaptation through experiences and collaborative efforts in multi-agent systems~\cite{dingDesignGPTMultiAgentCollaboration2023}. 
Reasoning is the process of processing and synthesizing information for problem-solving~\cite{park2023generative}. 
Decision-making in complex games demands multi-hop inference~\cite{li2023assessing} and long-term planning~\cite{hong2023metagpt}, incorporating sequential task decomposition and collaborative decision-making to effectively respond to dynamic games.
And reflection means the process of self-improvement, where agents evaluate and adjust their strategies based on feedback~\cite{wang2023voyager}. 
These components enable LMs-empowered agents to act effectively within the dynamic and evolving environments of modern digital games.

\noindent \textbf{Action} covers the interaction back with a game environment, i.e., operations executed by the agent as response to the game state and environment feedback. 
Behaviors are performed using generative coding, with techniques like iterative prompting~\cite{wang2023voyager}, role-specific engineering~\cite{dingDesignGPTMultiAgentCollaboration2023}, or code generations~\cite{wang2023voyager,tan2024towards}. 
Dialogue interactions span agent-agent and human-agent communications, employing collaborative frameworks~\cite{chen2023agentverse}, and conversation-driven controls for dynamic interactions~\cite{wu2023autogen}. 
Behavioral consistency in agents can be emphasized through, e.g., structural methods like Directed Acyclic Graph (DAG) for logical action progression~\cite{wu2023spring}, coupled with feedback mechanisms for environmental adaptation~\cite{anonymous2023ghost}, and reinforced by strategies like Reinforcement Learning (RL) for coherent action selection~\cite{xuLanguageAgentsReinforcement2023}.
These methodologies enable LMAs to not only handle complex tasks but also adapt behaviors to maintain consistency and alignment with game objectives in dynamic gaming contexts.

Nonetheless, challenges remain across all stages (and in other game scenarios). Four of them being especially significant in LMAs: i) addressing the problem of hallucinations in both critic agents and structured reasoning~\cite{hong2023metagpt, wu2023spring}; ii) correcting errors through iterative learning or feedback (e.g.,~\cite{li2023camel}); iii) generalizing learnt knowledge to unseen tasks, potentially using zero-shot learning or structured adaptability~\cite{wang2023voyager,zhang2023sprint}; and iv) interpretability, which requires transparent decision-making processes. 
While these are demonstrated in a diversity of AI systems, they also highlight the impacts of inherent limitations of LMs against the specific demands of gaming environments.

An overview of this survey structure is illustrated in Figure \ref{fig:taxonomy}. In \S\ref{sec:perception}, \S\ref{sec:brain}, and \S\ref{sec:action}, we review how existing LM-based game agents deal with perception, inference, and behavior. In \S\ref{sec:other_challenges}, we analyze the common challenges encountered throughout these three stages. Finally, in \S\ref{sec:Future Direction} we discuss future open research directions for furthering the development of general game-playing agents.

\section{Perception} \label{sec:perception}
Perceiving raw observations from the gaming world and extracting information are crucial for subsequent reflection and action selection. 
The primary function of perception is to transform the multi-modal space, encompassing vision (\S\ref{subsec:vision}), semantics (\S\ref{subsec:semantics}), and audio, into inputs for the agent. 
Text-based games primarily focus on communication and interaction among players, often through text or spoken words, to solve puzzles, reveal hidden information, or identify character roles, with games like Werewolf~\cite{xu2023exploring,xuLanguageAgentsReinforcement2023} and Avalon~\cite{wangAvalonGameThoughts2023,light2023from,lan2023llm} being quintessential examples. 
Digital games~\cite{wang2023voyager,openai2019dota}, by integrating multiple modality perception channels, offer richer and more immersive experiences, enabling deeper engagement with the game world.
However, unfortunately, not much effort in the existing literature has been dedicated to the integration of audio data into the training of LMs or to optimize game agents. This remains a topic for future exploration (\S\ref{sec:Future Direction}).

\subsection{Semantics} \label{subsec:semantics}
Traditionally, semantics-based perception of environmental elements (from locations to objects, characters to actions) in games relied mostly on text items and textual descriptions, encompassing a broad spectrum of descriptions, to natural language instructions and dialogues. 
In categorizing the text perception methods used by LMAs in gaming, we can distinguish them by the nature and complexity of textual inputs, ranging from basic text to advanced multi-modal input. This analysis delineates four primary classes: 
Basic text input, which involves simple user ideas or descriptions~\cite{li2023camel} and basic game state variables and dialogues~\cite{akoury2023framework, park2022social}. 
Structured or role-based inputs, characterized by the inclusion of character, story, role-related information~\cite{park2023generative, hong2023metagpt, wangRoleLLMBenchmarkingEliciting2023}, and skills~\cite{gong2023mindagent}. 
Environmental and context-specific inputs, including cases such as detailed game descriptors and contextual information~\cite{wu2023spring, zhang2023building}, and natural language instructions for tasks~\cite{zhang2023sprint, xu2023exploring}. 
Advanced multi-modal inputs, which integrate visual, audible, and textual data for decision-making~\cite{chen2023towards, yang2023octopus}, and the combination of dialogue, game state, and scripts for richer interaction~\cite{akoury2023towards}. 
This categorization underscores the adaptive capabilities of LMAs in processing a wide spectrum of semantic information, pivotal for enhancing interaction within gaming environments.

\subsection{Vision} \label{subsec:vision}
The initial efforts in LMAs showed promise, but primarily focused on translating the entire environmental experience into text by converting multimodal perception data into textual information through various models and APIs (e.g., \cite{wu2023visual}).
These methods, however, tend to ignore or lose complex visual information in this \emph{translation}, thus failing to accurately and intuitively reflecting the real environment (e.g., \cite{zheng2023steve}).
As such process can lose important information during the conversion from multi-modal to textual formats~\cite{wu2023smartplay,wang2023voyager}, it is likely to cause difficulty for agents to understand their situation or navigate effectively in target games, which emphasizes the need for more comprehensive sensory data processing in LMAs.
Addressing these impediments by leveraging the latest developments in Multi-modal Large Language Models (MLLMs), like ~\cite{gpt-4V}, becomes then a natural potential approach. By utilizing MLLMs, LMAs can obtain a richer perception of their surroundings, which facilitates more intricate cognitive processes and decision-making capabilities in multifaceted, real-world-like settings. 

The performance of various visual encoders, including both end-to-end trained and pre-trained models, in accurately capturing and learning from the unique visual aspects and gameplay dynamics of each game is evaluated in~\cite{schaferVisualEncodersDataEfficient2023}.
A video game bug detecting method~\cite{taesiri2024searching}, leverages CLIP for zero-shot learning, processes data by encoding video frames and text queries into embeddings, and calculates similarity scores to identify relevant problematic video sequences.
A benchmark called GlitchBench~\cite{taesiri2023glitchbench} evaluates LMM's ability to detect and interpret video game glitches by interpreting single-frame images.
Vision-based perception is also used for Success Detection and Scene Description~\cite{huangInner}.
Octopus~\cite{yang2023octopus} utilizes images as key training inputs for action planning. 
Steve-Eye~\cite{zheng2023steve} combines a visual encoder with a pre-trained LM for effective multi-modal interactions in open-world scenarios. 
Palm-e \cite{driess2023palm}, while not directly game-focused, also showcases the significance of visual inputs, using multi-modal sequences for tasks ranging from robotic manipulation to visual question answering.
Regarding this visual observation space, there are three distinct approaches for perception worth highlighting: i) Utilization of internal data representations via game-specific APIs to feed game-related information into LMAs (e.g., \cite{wang2023voyager}); ii) Model pre-training, based on the acquisition of image and action data (e.g., VPT~\cite{baker2022video}); and iii) Direct use of pure image information for processing and interpretation (e.g., Cradle~\cite{tan2024towards}).

% \subsection{Audio} \label{subsec:audio}

\section{Inference} \label{sec:brain}

Owing to its potential in supporting autonomy, reactivity, initiative, and sociability functionality~\cite{xi2023rise}, LMs are well positioned as core components of an intelligent agent's cognitive framework. 

Distinct requirements present themselves in different stages of gameplay though.

In the initial phase of a game, an agent is required to absorb essential common sense and game-specific background knowledge (either through pre-training or on the fly perception). 
In the course of the game, the agent's role extends to synthesizing past game events, managing knowledge storage and retrieval (\S\ref{subsec:memory}), and undertaking core cognitive functions such as information learning (\S\ref{subsec:learning}), reasoning (\S\ref{subsec:reason}), decision-making (\S\ref{subsec:decision}), and reflection (\S\ref{subsec:reflect}). Moreover, the agent continually updates or improves its knowledge base for future endeavors.

\subsection{Memory} \label{subsec:memory}
In order to properly represent learned knowledge or past event and use this information in inference, an agent need to effectively manipulate such \emph{memories}.
How to devise a high-quality memory mechanism that enables efficient retrieval and storage of memory by agents, while both making full use of backing LMs and respecting their constraints, remains a not-settled question.

\subsubsection{Common Sense.} Commonsense knowledge, typically acquired early in life and often unstated in a given context, is crucial for reasoning and avoiding misinterpretations. 
LMs, pre-trained on diverse internet corpora, exhibit advanced language comprehension and reasoning skills, using this implicit knowledge to adapt to new challenges. 
However, this knowledge might not always align with specific game scenarios. 
So current research aims to enhance LMs with this essential commonsense understanding in the following ways.
Embedding common sense using a structured format of goals and instructions is a key feature in \cite{anonymous2023ghost,zhang2023sprint}. 
A Standardized Operating Procedure (SOP) is utilized to integrate common sense into specific roles and tasks, showcasing an innovative approach to contextual understanding \cite{hong2023metagpt}.
Design thinking~\cite{dingDesignGPTMultiAgentCollaboration2023} and interpretation of robot action language instructions~\cite{xiao2022robotic} expand the common sense of LMAs.
Retrieval enhancement method~\cite{wangApolloOracleRetrievalAugmented2023} helps LMAs utilize external knowledge to break cognitive limitations.
These methods show that, with refined settings, LMs can grasp and apply common sense knowledge as intuitively as human understanding.

\subsubsection{Game Background Knowledge.} Game Background Knowledge in LMs refers to the understanding of various aspects of gaming environments. 
However integrating this knowledge with actual gameplay remains challenging~\cite{tsaiCanLargeLanguage2023,ma2023large}.
Current techniques enhance multi-agent systems with this knowledge by incorporating basic game rules, fundamental processes, reasoning skills, and one-shot demonstrations into prompts, ensuring a more comprehensive understanding of the game context. 
This includes: 
A benchmark called GameBugDescriptions~\cite{taesiri2022large} prompts LMs using zero-shot for background knowledge about a specific game by inputting game names.
The practice of inputting detailed character profiles into LMs to aid in simulating actions and dialogues is exemplified in  \cite{li2023camel,park2023generative,shao2023characterllm,wangRoleLLMBenchmarkingEliciting2023}.
Knowledge graphs are used to align player input with entities and relationships to provide game background knowledge \cite{ashby2023personalized}.
Furthermore, the incorporation of game-specific knowledge, such as poker strategies, enhances strategic decision-making capabilities \cite{gupta2023chatgpt}, indicating a focused approach towards improving models' understanding and functionality within specific contexts.
Integrating game background knowledge from datasets, such as Disco Elysium which includes dialogue and game state variables, enables LMs to understand and respond within the narrative context \cite{akoury2023towards}.
Incorporating narrative elements and structured game mechanics significantly improves LMs' contextual understanding \cite{akoury2023framework,zhu2023calypso}.
The ability of LMs to generate social behaviors based on community descriptions is illustrated by \cite{park2022social}.
Using detailed scenarios from simulated environments and Role-Playing Game (RPG) logs significantly enhances language models' understanding of game dynamics \cite{li2023metaagents,liang2023tachikuma}. 
Additionally, building an external knowledge base with documentation from the Minecraft Wiki and item crafting/smelting recipes creates an exhaustive knowledge source about the Minecraft world \cite{anonymous2023ghost}.
In summary, Game Background Knowledge in LMs encapsulates a broad spectrum of gaming-related information, ranging from basic rules and mechanics to complex narratives and role-specific knowledge, enabling these models to interact and perform more effectively in gaming environments.

\subsubsection{Retrieval.} Memory Retrieval in LMAs is a multifaceted process that enhances inference and actions by recalling learned information. 
In categorizing the memory retrieval methods used by LMAs in gaming, we can distinguish between structured memory systems, dynamic and adaptive retrieval processes, and advanced memory retrieval techniques. Structured memory systems, such as the skill library~\cite{wang2023voyager}, shared message pools~\cite{hong2023metagpt}, hierarchical structure \cite{liu2024llmpowered}, and structured communication~\cite{qian2023communicative}, focus on efficiently organizing and indexing information for quick retrieval. 
These systems allow agents to access previously acquired skills and knowledge, facilitating effective memory management. 
The "Summarize-and-Forget" memory mechanism efficiently prioritizes key memory items by summarizing important information and discarding the rest, reducing computational costs \cite{kaiya2023lyfe}.
The dynamic and adaptive retrieval process in~\cite{park2023generative} focuses on combining relevance, recency, and importance for memory retrieval, while strategic collection in~\cite{xu2023exploring} emphasizes the relevance and freshness of historical game information. This approach is enhanced with continuous feedback mechanisms for plan adjustments ~\cite{huangInner,wang2023describe}.
These methods demonstrate the ability of agents to dynamically adapt their memory retrieval to the context and ongoing inputs. 
Advanced memory retrieval techniques, such as the long-short term memory~\cite{zhou2023agents} and subtask re-labeling~\cite{feng2023llama}, generating concise state descriptions \cite{nottingham2023selective}, compressing historical text sequences \cite{piterbarg2024diff}, and the specialized memory retrieval modules for storing and accessing historical records~\cite{wu2023deciphering}, highlight the sophisticated approaches employed to manage memory in complex gaming scenarios. 
The use of a video-language model for task-relevant information encoding and retrieval further underscores the advanced capabilities of LMAs \cite{fan2022minedojo}. 
Collectively, these diverse methods illustrate the range and sophistication of memory retrieval strategies in LM-based game agents, showcasing their ability to efficiently store, access, and utilize information for effective and informed gameplay.

\subsection{Learning} \label{subsec:learning}
Learning in the context of LMs involves interpreting and integrating information from various sources, including training data, user interactions, and environmental feedback. 
The learning process can be explicit, such as through fine-tuning on specific datasets or task-focused training~\cite{feng2023llama}, or implicit, as seen in models leveraging pre-existing knowledge to adapt to new scenarios~\cite{huangInner,xu2023exploring}.
Iterative and Feedback-Based Learning~\cite{wang2023voyager,qian2023communicative,schaferVisualEncodersDataEfficient2023} merges iterative prompting, role specialization, feedback mechanisms, and behavior cloning, enhancing their capabilities through continuous feedback and adaptation.
Reinforcement and Offline Learning encompasses methods like RL with environmental feedback, offline learning, and language-instruction-conditioned RL, focusing on training LMs to make informed decisions based on environmental interactions and predefined instructions~\cite{yang2023octopus,zhang2023sprint,lan2023llm}.
Experience-Based and Memory Utilization Learning~\cite{xu2023exploring,ma2023large} focuses on leveraging past experiences, memory retrieval, and experience pools to inform current and future actions. 
Collaborative and Adaptive Learning~\cite{dingDesignGPTMultiAgentCollaboration2023,chen2023autoagents} emphasizes learning through multi-agent collaboration, adaptive strategies, and interactive learning processes, underlining the importance of collaborative problem-solving and adapting to new information or tasks. 
Simulated and Contextual Learning~\cite{park2022social,wu2023tidybot} pertains to learning within simulated environments or through context-based training, with models adapting to and learning from virtual environments using contextually rich scenarios. 
In summary, the diverse approaches to learning underscore a multi-faceted strategy, including iterative feedback, reinforcement techniques, experience-based strategies, collaborative processes simulated contexts, and so on.
These learning methodologies are increasingly tailored to leverage complex interactions and data integrations for enhanced cognitive and perceptual abilities in varied environments.

\subsection{Reasoning}\label{subsec:reason}
Reasoning involves abstracting the foundational knowledge acquired from the learning phase into more advanced forms of understanding through deductive, inductive, or abductive reasoning. This process encompasses the ability of LMs and AI agents to process information, draw conclusions, and make predictions; which can be involved in synthesizing data, understanding contexts, making inferences, and applying logical thinking to navigate complex situations or tasks. 
A strategic reasoning approach, engaging in abstract thinking and strategic planning, synthesizes complex information required for decision-making and extends reasoning beyond direct data \cite{park2023generative,wangAvalonGameThoughts2023}. 
Contextual and adaptive reasoning methods allow models to adjust their reasoning based on feedback, environmental factors, and changing scenarios, showcasing their adaptability through the process of model interpretation and adaptation to specific environments \cite{wang2023voyager,huangInner,tan2024towards}.
The utilization of multi-agent collaboration and interactive reasoning, where reasoning is a collective process involving various agents or factors working together to obtain a unified decision or solution \cite{hong2023metagpt,qian2023communicative}. 
Task-oriented reasoning that focuses on specific difficult problems and demonstrates directly  achieving established goals \cite{zhang2023sprint,anonymous2023ghost}. 
The exploration of reasoning within LMAs reflects a sophisticated blend of strategies that encompasses strategic, contextual, collaborative, and task-oriented approaches. These illustrate the capacity of LMAs to not only interpret and engage with complex data but also to dynamically adapt and respond to an ever-changing array of challenges and environments.

\subsection{Decision-making} \label{subsec:decision}
In the expansive and rich worlds of digital games, especially within the realms of open-world and immersive environments, players and AI agents are presented with complex decision and action spaces. Such spaces encapsulate the vast array of possible actions, interactions, and decisions available, characterized by an extensive variety of roles, item usage, and skill combinations. 
Moreover, multi-agent planning can add another layer of complexity, necessitating coordination and collaboration among multiple agents in environments with incomplete information, like in strategy games. Therefore, adaptation is key for agents to effectively interact in these environments, requiring a blend of strategic contextual problem-solving reasoning skills to effectively respond to dynamic and complex game worlds.
Such environments demand not only fast, reactive decision-making but also the capability to engage in multi-hop inference and long-term planning.

\subsubsection{Multi-Hop Inference.} Multi-hop inference refers to the cognitive process of considering multiple layers of information and dependencies before making a decision. This demands a high level of cognitive sophistication, where the agent must assess, predict, and integrate multiple strands of information to formulate strategies that are both effective and adaptable. For instance, in a strategy game, an agent must consider the current state of play, anticipate opponent moves, evaluate the potential outcomes of various actions, and then decide on the most advantageous course of action. 
Although LMs like GPT-4 have foundational decision-making abilities, they struggle with coherent multi-hop logical reasoning such as Minesweeper~\cite{li2023assessing}.
The involvement of multiple agents working together enhances multi-step decision-making through knowledge sharing \cite{hong2023metagpt,qian2023communicative,chen2023agentverse}. 
Multi-level decision-making in a hierarchical structure for detailed and strategic choices is adopted by \cite{yuan2023skill,zhang2023proagent,cai2023groot}.
Automated process and SOP management~\cite{zhou2023agents,wang2023describe} use standardized procedures to make consistent decisions.
Although these strategies underscore the ongoing efforts to enhance LMAs' multi-hop inference capabilities, the transition to coherent multi-hop inference in complex scenarios remains a work in progress.

\subsubsection{Long-Term Planning.} Long-term planning is the strategic process of setting goals and determining actions to achieve these goals over an extended period, involving multiple steps or stages.
In gaming, this could translate to developing a strategy, requiring the player or LMA to make decisions that not only have immediate benefits but also contribute to their overarching objectives. 
In the realm of LMAs tackling long-term planning challenges in gaming contexts, a broad categorization reveals several key approaches. 
Chain-of-Thought (COT)~\cite{wei2022chain} planning generates intermediate steps for a final decision, mimicking human reasoning where each step builds on the previous for a conclusive outcome~\cite{wu2023visual}.
Complex tasks are decomposed into manageable sub-tasks, emphasizing step-by-step decision-making \cite{dingDesignGPTMultiAgentCollaboration2023,jin2023alphablock}. 
Structured planning and iterative processes~\cite{hong2023metagpt,qian2023communicative,chen2023agentverse} emphasize the importance of organized workflows and the iterative refinement of strategies. 
Skill pre-training and adaptive learning~\cite{zhang2023sprint,zhaiBuildingOpenEndedEmbodied2023} underscores the necessity for agents to be equipped with a diverse set of pre-trained skills that enable them to adapt to new scenarios efficiently.
Finally, these approaches to long-term planning in the context of LMAs within gaming illustrate a strategic convergence of methodologies aimed at enhancing decision-making and problem-solving capabilities.
Through CoT planning, structured and iterative processes, as well as skill pre-training and adaptive learning, LMAs are being progressively refined to tackle complex, long-term planning challenges. 

\subsection{Reflection.} \label{subsec:reflect} Reflection refers to the capacity of LMAs to evaluate, assess, and potentially adjust their own processes, decisions, or outputs as tasks progress. 
It implies a level of meta-cognition where the system considers its actions, their outcomes, and the context in which they occur. 
Reflection can lead to improvements in performance, the refinement of strategies, or the reassessment of goals.
Feedback-Based Self-Improvement~\cite{wang2023voyager,anonymous2023ghost,chen2023agentverse,hong2023metagpt,lan2023llm} includes processes that utilize external feedback, such as environmental responses or code execution outcomes, for self-verification, analysis, and iterative enhancement. 
Iterative Optimization of Plans and Actions~\cite{huangInner,zhaiBuildingOpenEndedEmbodied2023,cai2023groot} focuses on methods that continuously refine plans or actions based on feedback and new information to better achieve or adapt to goals. 
Collaboration and Multi-Agent Interaction~\cite{qian2023communicative,wangAvalonGameThoughts2023} emphasizes collaborative reassessment and adjustment of decisions in multi-agent systems. 
Debate and Theory of Mind-Based Reflection~\cite{duImprovingFactualityReasoning2023,guo2023suspicion} focus on models that employ debate or exchange formats as reflective processes. 
Thus, reflection embodies a critical aspect of LMAs, enabling them to evaluate and refine their processes through feedback-based self-improvement, iterative optimization, collaborative interactions, and debate-driven reflection. This capacity for self-assessment and adjustment is instrumental in enhancing decision-making, strategy development, and goal reassessment, ultimately leading to more sophisticated, self-aware, and adaptable solutions.

\section{Action} \label{sec:action}
In this section, we will explore how LMs exhibit human-like action in-game environments, including specific behavior execution (\S\ref{subsec:behave}), communication with humans or other agents (\S\ref{subsec:dialogue}), and how to ensure that these actions are consistent (\S\ref{subsec:consistent}). 
These agents utilize generative programming techniques, interactive feedback with the environment, and complex conversational exchanges with other agents or human players to perform tasks and solve in-game challenges.

The action spaces within which LMAs operate in-game can be broadly categorized into three distinct types, each with its unique challenges and opportunities for interaction and control. These categories range from purely linguistic engagements to direct manipulation of game controls, not only defining the scope of actions available to LMAs but also shaping the strategies and technologies employed to navigate them. \textbf{i) Text-Based Interaction.} 
The first category encompasses pure linguistic interaction,  which primarily focuses on linguistic communication and interaction among players. These games, such as Werewolf (e.g., \cite{xu2023exploring}) and Avalon (e.g., \cite{wangAvalonGameThoughts2023}), revolve around dialogue, decision-making, and the interpretation of textual information. In these environments, LMAs are required to understand and generate natural language, engaging with players and the game's narrative through text. This demands a deep understanding of language nuances, player intentions, and the ability to craft responses that can influence the game's outcome. 
\textbf{ii) API or Predefined Actions.}
The second category involves manipulating the game environment through APIs or pre-defined actions, offering a more structured approach to game mechanics.
Examples include the use of the Mineflayer JavaScript APIs in Voyager for action controls \cite{wang2023voyager} and GITM opting for structured actions implemented via hand-written scripts \cite{anonymous2023ghost}. This method requires understanding of the game's mechanics and the ability to strategically select and sequence actions to achieve desired outcomes, but benefits from the extra semantics and control provided by game-specific APIs.
\textbf{iii) Direct Control via IO Operations.}
The third category represents the most immersive form of interaction: direct control only over input devices, like mouse and keyboard. VPT \cite{baker2022video} and Cradle \cite{tan2024towards} operate using the same IO devices available to users, e.g., mouse and keyboard, and their input space is discussed in visual observations  (\S\ref{subsec:vision}). 
This approach simulates the human gaming experience most closely, with LMAs performing actions at a higher abstraction level, navigating menus, and manipulating items just as a human player would.
This represents a more general form of interaction, necessitating a significant integration of cognitive processing and motor skills.
Each of these categories showcases the versatility and potential of LMAs in gaming, from the purely cognitive and linguistic challenges of text-based games to the physical and tactical demands of direct control.

\subsection{Behavior} \label{subsec:behave}

LMAs perform specific actions through generative programming techniques.
Iterative prompting that incorporates environmental feedback, execution errors, and self-verification to continuously refine generated programs, ensuring their effectiveness and relevance to tasks at hand \cite{wang2023voyager}. 
Programs are generated by interpreting existing instructions or employing role-specific prompt engineering \cite{zhang2023sprint,dingDesignGPTMultiAgentCollaboration2023}, to enhance the specificity of tasks addressable by LMAs. 
Generating programs (task plans) by interpreting natural language instructions within the context of a 3D Scene Graph, which maps instructions to actionable steps for robot execution \cite{rana2023sayplan,wu2023embodied}.
By continuously refining their outputs through feedback loops and leveraging detailed environmental data, LMAs demonstrate an evolving capacity for precise and contextually relevant action planning.

\subsection{Dialogue} \label{subsec:dialogue}
LMAs interact with humans or other agents by taking into account the current environment and past experiences.
To categorize the described interaction methods, we can distinguish them based on their primary focus: agent-agent interactions, and human-agent interactions. 
Some methods are versatile, covering both categories. 
The collaboration framework of LMAs is divided into two methods: horizontal communication (mutual equality) and vertical communication (there is a superior-subordinate relationship between agents or between humans and agents) \cite{chen2023agentverse}.
Agent-agent interactions include collaborative frameworks~\cite{chen2023autoagents}, conversation-driven control in multi-agent systems~\cite{wu2023autogen, zhou2023agents, fuImprovingLanguageModel2023}, and game-based strategic behaviors~\cite{xu2023exploring, light2023from,abdelnabi2023llmdeliberation}.
On the other hand, human-agent interactions are characterized by instruction-following and collaborative task solving~\cite{li2023camel, park2023generative, qian2023communicative}, gameplay enhancement and creative assistance~\cite{akoury2023framework, zhu2023calypso, dingDesignGPTMultiAgentCollaboration2023}, natural and flexible communication~\cite{yang2023octopus}.
Additionally, Lyfe Agents \cite{kaiya2023lyfe} exemplify autonomy within human-agent and agent-agent interactions by navigating to various locations, forming groups, and engaging in proximity-based conversations, which includes autonomously choosing to ignore or leave conversations, showcasing a practical application of agent autonomy and social behaviors.
Additionally, LMAs exhibit human-like social behaviors in collaboration, such as Teamwork, Leadership, Persuasion, Deception, and Confrontation~\cite{lan2023llm}.
This multifaceted approach facilitates a spectrum of interactions, from cooperative frameworks to strategic gameplay and creative task-solving, underpinning the importance of versatile, strategic, and socially aware communication methods in the advancement of LMAs.

\subsection{Consistency} \label{subsec:consistent}

Action consistency ensures that agent actions remain coherent and aligned with their objectives.
This consistency is maintained through consideration of past experiences and current context, utilization of feedback mechanisms, and modification of actions and plans when necessary.
Specific strategies for ensuring consistent actions include look-ahead checking mechanisms to assess action feasibility~\cite{gong2023mindagent}, leveraging visual feedback mechanisms for task-specific responses~\cite{wu2023visual}, using RL to train policies that select the most appropriate actions~\cite{xuLanguageAgentsReinforcement2023}, and employing the Self-Consistency (SC) \cite{wang2023selfconsistency} such as PokéLLMon~\cite{hu2024pokellmon} which generates multiple action predictions and selects the most common one for consistent decision-making.
Structural approaches include using DAG and game-related structures for logical action progression~\cite{wu2023spring}, employing skill libraries for consistent behavior execution~\cite{wang2023voyager}, and defining structured actions with clear semantics~\cite{anonymous2023ghost}. 
Furthermore, assigning specific roles to agents and following structured workflows helps maintain consistency by aligning outputs with defined roles and tasks~\cite{hong2023metagpt}, while SOP offers fine-grained control over behavior~\cite{zhou2023agents}. 
Feedback and adaptation mechanisms play a crucial role, with techniques ranging from revising plans based on environmental feedback~\cite{feng2023llama} to employing unified interfaces for automatic responses~\cite{wu2023autogen}. 
Neuro-Symbolic Approach (e.g. COTTAGE \cite{dong2023cottage}) introduced addresses the consistency issue in text-based adventure games generated by LMs. This approach combines reasoning and code generation for dynamic game state tracking and improved player interaction, ensuring narrative coherence and continuity.
Through these mechanisms, large model agents can not only perform complex and diverse tasks in the game, but also engage in highly interactive dialogue exchanges with human players or other agents, while ensuring the consistency and adaptability of their actions.

\begin{figure*}
    \centering
    \includegraphics[width=1.05\linewidth]{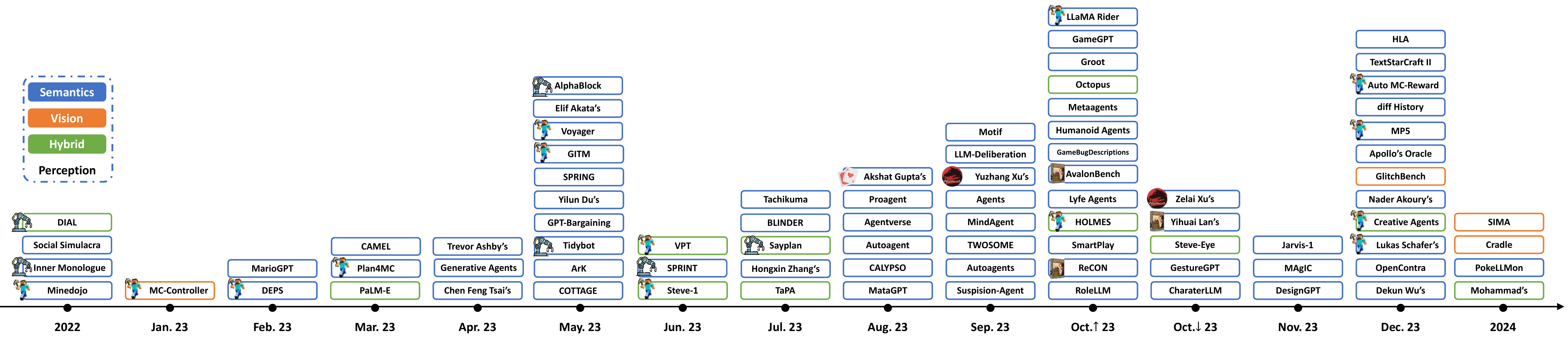}  
    \caption{Overview of Representative Works on Game-Playing Agents and Large Models.} 
    \label{fig:timeline}
\end{figure*}

\section{Challenges Throughout the Playing Cycle} \label{sec:other_challenges}
In exploring the gameplay of LMAs, several key challenges emerge, including resolving hallucinations (\S\ref{subsec:hallu}), error correction (\S\ref{subsec:error}), generalizing to unseen tasks (\S\ref{subsec:gene}), and enhancing interpretability (\S\ref{subsec:inter}). While existing research proposes some solutions, these challenges remain significant obstacles for LMAs in gaming contexts.

\subsection{Hallucination} \label{subsec:hallu}
LMs can produce outputs that diverge from original or accurate information, a situation commonly described as hallucinations.
Addressing hallucinations is complex, with LMAs sometimes sending messages with grounding errors, contradictions, or strategic subpar content \cite{HumanlevelPlayGame2022}. 
To classify the various methods employed for solving hallucinations in LMAs, we can look at strategies ranging from structured reasoning and contextual awareness to interactive and collaborative methods, as well as specific prompt or feedback mechanisms. 
Studies introduce critic agents, SOP, or role-specific prompts to reduce logical hallucinations, as shown in MetaGPT and JARVIS-1~\cite{hong2023metagpt,wang2023jarvis}. 
Structured reasoning frameworks like DAG used in~\cite{wu2023spring} provide a solid foundation for maintaining coherence. 
A multi-faceted approach \cite{chen2023gamegpt} is employed that encompasses dual collaboration, code decoupling, the development of in-house lexicons, and direct interaction with users for plan rectification.
Additionally, breaking down tasks into smaller components and employing cross-examination in~\cite{qian2023communicative}, or using role-specific cues and character context as seen in~\cite{xu2023exploring}, effectively mitigate hallucinations. 
The interactive and collaborative aspect includes allowing for human editing of AI-generated content \cite{zhu2023calypso}, and utilizing structured multi-agent collaboration and feedback mechanisms \cite{chen2023agentverse}.
Pok\'eLLMon~\cite{hu2024pokellmon} addresses the hallucination problem by integrating In-context Reinforcement Learning (ICRL) and Knowledge-Augmented Generation (KAG) to improve decision-making accuracy and leverage external knowledge for informed actions.
Visual fidelity to outputs from Visual Foundation Models (VFMs) is another approach~\cite{wu2023visual}. 
The incorporation of external knowledge sources in~\cite{wangApolloOracleRetrievalAugmented2023} further enhances the accuracy of LMAs-generated content. 
Specific prompt or feedback mechanisms like reducing contradictory messages~\cite{HumanlevelPlayGame2022}, prompt reconstruction~\cite{dingDesignGPTMultiAgentCollaboration2023}, confidence scoring~\cite{wangRoleLLMBenchmarkingEliciting2023}, and regular expression matching in MindAgent~\cite{gong2023mindagent} contribute to reducing inaccuracies.
Debates are used to refine responses~\cite{duImprovingFactualityReasoning2023}, and task-specific fine-tuning is employed in~\cite{jin2023alphablock}. 
Finally, TaPA \cite{wu2023embodied} ensures that generated action plans only involve objects present in the scene to maintain accuracy. 
While diverse methods like structured reasoning, interactive editing, and specific mechanisms play a crucial role in mitigating hallucinations in AI systems and enhancing the coherence and accuracy of outputs, they do not entirely eliminate the issue of hallucinations.

\subsection{Error Correction}\label{subsec:error}
In terms of error correction, various studies employ iterative prompting mechanisms, executable feedback mechanisms, or chat chains among models to identify and correct errors.
The use of critic agents for selection and feedback~\cite{li2023camel}, iterative prompting with environment feedback~\cite{wang2023voyager}, and executable feedback for debugging~\cite{hong2023metagpt} highlights the importance of iterative learning and feedback mechanisms in refining actions and decisions.
Iterative re-planning based on real-time feedback is employed in~\cite{rana2023sayplan, wang2023jarvis} while focusing on learning from mistakes to improve game strategies is adapted by ~\cite{lan2023llm}. 
MindAgent \cite{gong2023mindagent} through an Action Validation mechanism that employs a look-ahead checking system to parse and assess the feasibility of proposed actions, returning an error message if an action is found to be inexecutable.
Auto MC-Reward \cite{li2023auto} evaluates and corrects reward functions for accuracy, consistency, and relevance, including managing runtime errors with iterative feedback up to three times.
Peer review and system testing~\cite{qian2023communicative}, evaluation stage feedback~\cite{chen2023agentverse}, and the use of an additional model, MarioBert~\cite{sudhakaranMarioGPTOpenEndedText2Level2023}, represent typical approaches to error resolution.
Structured systems and processes, such as the historical experiences and reflections in~\cite{xu2023exploring}, debate structures and evidence pools in~\cite{wangApolloOracleRetrievalAugmented2023}, and the debate process for refining responses in~\cite{duImprovingFactualityReasoning2023}, contribute to minimizing errors. 
Specific strategy implementations like the trial-and-error approach in~\cite{yang2023octopus} and the closed-loop approach with visual feedback in~\cite{jin2023alphablock} are vital for correcting errors and enhancing accuracy. 
While diverse methods such as iterative feedback, specialized tools, structured processes, and strategic implementations are crucial for effective error correction in LMAs and improve agent adaptability, there is still a need to enhance the efficiency and accuracy of these processes.

\subsection{Generalization}\label{subsec:gene}
The ability to generalize involves agents applying the knowledge, skills, and abilities gained in one context to achieve goals and perform tasks in new, previously unencountered environments, allowing for ongoing adaptation and development.
The method of COT reasoning guided by context~\cite{wu2023spring}, along with the GITM's utilization of long-term planning and external knowledge in Minecraft~\cite{anonymous2023ghost}, and the emphasis on information extraction and collaborative planning in multi-agent settings~\cite{zhang2023building}, highlight contributions to generalizing to unseen tasks in LMAs. 
VOYAGER \cite{wang2023voyager} and the DEPS \cite{wang2023describe} showcase the ability to handle new tasks without prior specific training, demonstrating zero-shot generalization, while RoleLLM exhibits robust generalization to unseen roles with minimal input~\cite{wangRoleLLMBenchmarkingEliciting2023}. 
ChatDev's discussion of adaptive software development using the chat chain framework ~\cite{qian2023communicative} and AlphaBlock's structured framework for adaptability in creating subtask plans for various robot operation scenarios~\cite{jin2023alphablock} is highlighted.
Pre-training based on language instructions~\cite{zhang2023sprint} demonstrates effective fine-tuning and the ability to transfer skills to new environments, outperforming other methods in learning tasks in new environments.
Cradle \cite{tan2024towards} and SIMA \cite{sima} take different approaches to extend LMA generalization to easily adapt to new environments. Both use a common human-like interface to interact with the game environment, where  input consists of image observations and language instructions, and output takes the form of keyboard and mouse operations. While SIMA focuses on pre-training a new model, Cradle focuses on leveraging LMs capabilities into a cognitive agent framework.
These methods showcase the capabilities of LMAs in adapting to unfamiliar tasks while acknowledging that generalization capabilities in more complex or variable environments continue to present significant challenges.

\subsection{Interpretability}\label{subsec:inter}
Regarding interpretability, some research has made strides in enhancing the explainability of agent actions through COT processes, modular decision-making processes, or clear conversational programming paradigms. However, the logic behind agents' decisions and behaviors remains challenging to fully comprehend, especially in complex gaming environments.
For reasoning and decision-making process transparency, SPRING~\cite{wu2023spring} and LLaMA Rider~\cite{feng2023llama} employ COT reasoning to make actions understandable, while Generative Agents~\cite{park2023generative} bases explainability on a combination of past experiences and reflections, and AlphaBlock~\cite{jin2023alphablock} derives explainability from logical sub-task plans based on instructions and observations. 
For structured and modular approaches, Voyager~\cite{wang2023voyager} utilizes a skill library and iterative prompting for transparent decision-making, GITM~\cite{anonymous2023ghost} focuses on structured actions with clear semantics for explainability, and a modular approach in planning and decision-making enhances understandability~\cite{zhang2023building}. 
Structured communication and role-based task management MetaGPT \cite{hong2023metagpt} and Agentverse \cite{chen2023agentverse} further aid in making agent actions interpretable. 
Action explanation~\cite{qian2023communicative} includes a chat chain offering a transparent view of the software development process, enhancing explainability through structured conversation programming~\cite{wu2023autogen}, and integrating cognitive processes for explainability in deceptive environments~\cite{wangAvalonGameThoughts2023}. 
Additionally, DEPS~\cite{wang2023describe} uses an LLM-based explainer for plan failures, Grad-CAM is employed for visual attention interpretability~\cite{schaferVisualEncodersDataEfficient2023}, and the decision-making process is made more transparent through detailed prompts guiding the use of VFMs~\cite{wu2023visual}. 
The clear role definition in debates~\cite{wangApolloOracleRetrievalAugmented2023}, RL and diverse action candidates~\cite{xuLanguageAgentsReinforcement2023}, imitation learning and expert behavior mimicking~\cite{huang2023ark}, and behavior cloning and networks for explicit action choices~\cite{cai2023groot}, all contribute to making agent actions more explainable. 
These diverse methods make the decision-making of LMAs more understandable and traceable.

\section{Conclusion and Future Direction}\label{sec:Future Direction}
In this paper, we survey the literature on combining agents and LMs towards playing complex digital games  (\autoref{fig:timeline}), which to our knowledge is the first review to thoroughly investigate game-playing by LMAs.
We provide a detailed overview of the various challenges LMAs face in open gaming worlds.
By emphasizing the typical agent workflow - perception, reasoning, and behavior - we can compare and contrast how LMs can be applied to different problems at different stages and elaborate on specific solutions of individual LM-empowered game agent approaches.

Besides providing an overview of current approaches and common challenges, based on the executed literature review this work also identifies future research directions for LMAs and digital games.

\textbf{Multi-modal Perception:}
Despite advancements, LMAs like VOYAGER still lack visual perception due to the unavailability of models until recently, suggesting a significant area for improvement~\cite{wang2023voyager}. 
Furthermore, sound effects in games, are crucial for perceiving context, understanding tasks, and receiving feedback—key components for the successful completion of game objectives. Despite their significance, there is a scarcity of literature exploring how auditory information can be leveraged to enhance game agents' performance.
Enhancing multi-modal capabilities, including visual and auditory perceptions, could lead to more sophisticated task handling and immersive game playing experiences.

\textbf{Authenticity in Gaming Experience:}
The evolution towards authenticity in gaming scenarios also merits further exploration. 
Studies have shown a preference for human-written content over LLM-generated dialogue and ideas, indicating the need for better grounding of LLM generations in the game's narrative and state~\cite{park2023generative,akoury2023framework, zhu2023calypso,wang2023open}. Enhancements in narrative generation and character simulation, as explored in RoleLLM, could lead to more realistic and immersive interactions~\cite{wangRoleLLMBenchmarkingEliciting2023}. Furthermore, the behaviour itself of LMAs is oftentimes artificial and making it more fluid/realistic opens new possibilities.

\textbf{Use of External Tools:}
LMA's inability to master leveraging external tools to enhance gameplay represents another significant gap in achieving AGI.
Current LMs struggle with selecting and employing tools efficiently, resulting in limited access to real-time information. 
Enabling the ability to dynamically access and interpret online game guides and generalize external game-specific knowledge to new games will represent significant progress, and this will require enhanced model capabilities to interact with external APIs and documentation.
Works such as Toolformer~\cite{schick2023toolformer} and Gorilla~\cite{patil2023gorilla} address some of these challenges by focusing on re-representing API call primitives and fine-tuning models to better understand and use documents. However, a full implementation of LM game agents proficient in the use of external tools remains open.

\textbf{Real-Time Gaming:}
Considering the inherent reasoning process and computational demands of LMs, mastering real-time, high-paced gaming presents a formidable challenge. The demanding nature of gaming environments necessitates adherence to rigorous real-time performance and inference speed criteria. While LMs offer considerable advantages, they are not a substitute for RL in contexts requiring immediate, time-critical decision-making \cite{iskander2020reinforcement,wei2022honor,openai2019dota}. Looking ahead to the integration of LMAs in live gaming sessions alongside human participants, these agents must achieve latencies aligned with the required frame rates to guarantee real-time interaction. Therefore, the pursuit of greatly improving the efficiency and reaction times of LMAs in dynamic and swiftly evolving gaming landscapes is essential for forthcoming progress.

In conclusion, while LMAs have made remarkable strides in gaming, their full potential is yet to be realized. Future work should focus on improving multi-modal perception, achieving greater authenticity in gaming experiences, effectively integrating external tools, and excelling in real-time gaming environments; among other unexplored fronts (e.g., LMAs as automatic game testers). These advancements will not only enhance the capabilities of LMAs, lead to more engaging and realistic gaming experiences but also lead to significant impact on real-world scenarios.
Moreover LMs can also be applied in diverse other digital game-related scenarios, e.g., level and story generation, or game balancing and mechanics tuning; which were outside the scope of this survey and deserve their own in-depth research.

\bibliographystyle{named}
\bibliography{ijcai24}
\end{document}

%% file: framework_setting.tex
\usepackage{natbib}
\setcitestyle{numbers,square}

\usepackage[utf8]{inputenc} % allow utf-8 input
\usepackage[T1]{fontenc}    % use 8-bit T1 fonts
\usepackage{url}            % simple URL typesetting
\usepackage{color, soul}

\usepackage{booktabs}       % professional-quality tables
\usepackage{amsfonts}       % blackboard math symbols
\usepackage{nicefrac}       % compact symbols for 1/2, etc.
\usepackage{microtype}      % microtypography
\usepackage[dvipsnames]{xcolor}         % colors
\usepackage{graphicx}
\usepackage{csquotes}
\usepackage{amsmath}
\usepackage{enumitem}       
\usepackage{blindtext} %This package generates automatic text
\usepackage{epigraph} 

\usepackage{adjustbox}
\usepackage{tikz}
\usepackage{pgfplots}
\pgfplotsset{compat=1.18}
\usepackage[edges]{forest}
\definecolor{framework-blue}{RGB}{47, 85, 151}

\definecolor{content-yellow}{RGB}{255, 230, 153}
\definecolor{framework-yellow}{RGB}{255, 255, 255}
\definecolor{content-orange}{RGB}{251, 229, 215}
\definecolor{framework-orange}{RGB}{248, 203, 175}
\definecolor{content-gray}{RGB}{237, 237, 237}
\definecolor{framework-gray}{RGB}{166, 166, 166}

\definecolor{paired-light-blue}{RGB}{198, 219, 239}
\definecolor{paired-dark-blue}{RGB}{49, 130, 188}
\definecolor{paired-light-orange}{RGB}{251, 208, 162}
\definecolor{paired-dark-orange}{RGB}{230, 85, 12}
\definecolor{paired-light-green}{RGB}{199, 233, 193}
\definecolor{paired-dark-green}{RGB}{49, 163, 83}
\definecolor{paired-light-purple}{RGB}{218, 218, 235}
\definecolor{paired-dark-purple}{RGB}{117, 107, 176}
\definecolor{paired-light-gray}{RGB}{217, 217, 217}
\definecolor{paired-dark-gray}{RGB}{99, 99, 99}
\definecolor{paired-light-pink}{RGB}{222, 158, 214}
\definecolor{paired-dark-pink}{RGB}{123, 65, 115}
\definecolor{paired-light-red}{RGB}{231, 150, 156}
\definecolor{paired-dark-red}{RGB}{131, 60, 56}
\definecolor{paired-light-yellow}{RGB}{231, 204, 149}
\definecolor{paired-dark-yellow}{RGB}{141, 109, 49}
\tikzset{%
    parent/.style = {align=center,text width=2.5cm,rounded corners=3pt, line width=0.3mm, fill=gray!10,draw=gray!80},
    child/.style = {align=center,text width=2.3cm,rounded corners=3pt, fill=blue!10,draw=blue!80,line width=0.3mm},
    grandchild/.style = {align=center,text width=2cm,rounded corners=3pt},
    greatgrandchild/.style = {align=center,text width=1.5cm,rounded corners=3pt},
    greatgrandchild2/.style = {align=center,text width=1.5cm,rounded corners=3pt},    
    referenceblock/.style =  {align=center,text width=1.5cm,rounded corners=2pt},
    brain/.style = {align=center,text width=2.2cm,rounded corners=3pt, fill=white,draw=framework-blue,line width=0.3mm},   
    brain_work/.style = {align=center, text width=4.5cm,rounded corners=3pt, fill=white,draw=framework-blue,line width=0.3mm},
    perception/.style= {align=center,text width=2.2cm,rounded corners=3pt, fill=white,draw=framework-blue,line width=0.3mm},
    perception_work/.style= {align=center, text width=4.5cm,rounded corners=3pt, fill=white,draw=framework-blue,line width=0.3mm}, 
    action/.style= {align=center,text width=2.2cm,rounded corners=3pt, fill=white,draw=framework-blue,line width=0.3mm},
    action_work/.style= {align=center, text width=4.5cm,rounded corners=3pt, fill=white,draw=framework-blue,line width=0.3mm},
    single_agent/.style= {align=center,text width=2.2cm,rounded corners=3pt, fill=white,draw=framework-blue,line width=0.3mm},
    single_agent_work/.style= {align=center, text width=4.5cm,rounded corners=3pt, fill=white,draw=framework-blue,line width=0.3mm},
    multi_agent/.style= {align=center,text width=2.2cm,rounded corners=3pt, fill=white,draw=framework-blue,line width=0.3mm},
    multi_agent_work/.style= {align=center, text width=4.5cm,rounded corners=3pt, fill=white,draw=framework-blue,line width=0.3mm},
    human_agent/.style= {align=center,text width=2.2cm,rounded corners=3pt, fill=white,draw=framework-blue,line width=0.3mm},
    human_agent_work/.style= {align=center, text width=4.5cm,rounded corners=3pt, fill=white,draw=framework-blue,line width=0.3mm},
    behavior_and_personality/.style= {align=center,text width=2.2cm,rounded corners=3pt, fill=white,draw=framework-blue,line width=0.3mm},
    behavior_and_personality_work/.style= {align=center, text width=4.5cm,rounded corners=3pt, fill=white,draw=framework-blue,line width=0.3mm},
    society_environment/.style= {align=center,text width=2.2cm,rounded corners=3pt, fill=white,draw=framework-blue,line width=0.3mm},
    society_environment_work/.style= {align=center, text width=4.5cm,rounded corners=3pt, fill=white,draw=framework-blue,line width=0.3mm},
    society_simulation/.style= {align=center,text width=2.2cm,rounded corners=3pt, fill=white,draw=framework-blue,line width=0.3mm},
    society_simulation_work/.style= {align=center, text width=4.5cm,rounded corners=3pt, fill=white,draw=framework-blue,line width=0.3mm},
}

%% file: tree.tex
\newcommand{\textwidthtitle}{2cm}
\newcommand{\miniheighttitle}{10mm}
\newcommand{\filltitle}{white!20}

\newcommand{\textwidthsection}{2.2cm}
\newcommand{\miniheightsection}{8mm}
\newcommand{\fillsection}{white!10}

\newcommand{\textwidthsubsection}{2.1cm}
\newcommand{\miniheightsubsection}{6mm}
\newcommand{\fillsubsection}{cyan!10}

\newcommand{\textwidthsubsubsection}{7.2cm}
\newcommand{\textwidthsubsubsectionrealtitle}{2cm}
\newcommand{\fillsubsubsection}{yellow!15}

\newcommand{\textwidthexample}{6.3cm}
\newcommand{\fillexample}{yellow!15}
\newcommand{\keyexample}{Key Example}

\begin{figure*}[!t]
	\tiny
	\centering
	\begin{forest}
		for tree={
			forked edges,
			grow'=0,
			draw,
			rounded corners,
			node options={align=center},
		},
		[\textbf{Game Playing and Large Models}, text width=\textwidthtitle, minimum height=\miniheighttitle, fill=\filltitle
			[\textbf{Perception} \S\ref{sec:perception}, text width=\textwidthsection, minimum height=\miniheightsection, for tree={fill=\fillsection}
				[\textbf{Semantics}  \S\ref{subsec:semantics}, text width=\textwidthsubsection, for tree={fill=\fillsubsubsection}, minimum height=\miniheightsubsection, fill=\fillsubsection
					[
					{Basic Text Inputs~\cite{li2023camel,akoury2023framework, park2022social},
                    Structured or Role-based Inputs~\cite{park2023generative, hong2023metagpt, wangRoleLLMBenchmarkingEliciting2023,gong2023mindagent},\\
                    Environmental and Context-specific Inputs~\cite{wu2023spring, zhang2023building,zhang2023sprint, xu2023exploring},\\
                    Advanced Multi-modal Inputs \cite{chen2023towards, yang2023octopus,akoury2023towards}},
					text width=\textwidthsubsubsection,
                    rounded corners=false,
					node options={align=left}
					]
				]
                [\textbf{Vision} \S\ref{subsec:vision}, text width=\textwidthsubsection, for tree={fill=\fillsubsubsection}, minimum height=\miniheightsubsection, fill=\fillsubsection
                    [
					{Game Data API Integration (e.g., \cite{wang2023voyager}),\\
                    Image Data Pre-training (e.g., \cite{baker2022video}),
                    Pure Image Information Processing (e.g., \cite{tan2024towards})},
					text width=\textwidthsubsubsection,
                    rounded corners=false,
					node options={align=left}
					]
                ]
			]
            [\textbf{Inference} \S\ref{sec:brain}, text width=\textwidthsection, minimum height=\miniheightsection, for tree={fill=\fillsection}
				[\textbf{Memory} \S\ref{subsec:memory}, text width=\textwidthsubsection, for tree={fill=\fillsubsubsection}, minimum height=\miniheightsubsection, fill=\fillsubsection
					[
					\textbf{Common Sence},
					text width=\textwidthsubsubsectionrealtitle,
					node options={align=center}
					    [{Embedding with Structured Formats \cite{anonymous2023ghost,zhang2023sprint},\\
                        Standard Operating Procedures \cite{hong2023metagpt},
                        Design Thinking~\cite{dingDesignGPTMultiAgentCollaboration2023},\\
                        Interpreting Action Instructions~\cite{xiao2022robotic},\\
                        Retrieval Enhancement Methods~\cite{wangApolloOracleRetrievalAugmented2023}
}, 
					    text width=\textwidthexample, 
					    node options={align=left}, 
					    rounded corners=false,
					    fill=\fillexample]
					]
                    [
                    \textbf{Game Background Knowledge},
					text width=\textwidthsubsubsectionrealtitle,
					node options={align=center}
					    [{
                        Character Simulation and Dialogue Generation \cite{li2023camel,park2023generative,shao2023characterllm,wangRoleLLMBenchmarkingEliciting2023},\\
                        Incorporation of Narrative Elements and Game Mechanics \cite{akoury2023framework,zhu2023calypso},\\
                        Game Background Knowledge Integration \cite{taesiri2022large,gupta2023chatgpt,akoury2023towards},\\
                        Generation of Social Behaviors Based on Community Descriptions  \cite{park2022social},\\
                        Utilization of Simulated Environments and RPG Logs \cite{li2023metaagents,liang2023tachikuma},\\
                        Building an External Knowledge Base \cite{anonymous2023ghost,ashby2023personalized}
                        }, 
					    text width=\textwidthexample, 
					    node options={align=left}, 
					    rounded corners=false,
					    fill=\fillexample]
                    ]
                    [
                    \textbf{Retrieval},
					text width=\textwidthsubsubsectionrealtitle,
					node options={align=center}
					    [{Structured Memory Systems \cite{wang2023voyager,hong2023metagpt,liu2024llmpowered,qian2023communicative},\\
                        Dynamic and Adaptive Retrieval Processes \cite{kaiya2023lyfe,park2023generative,xu2023exploring,huangInner,wang2023describe},\\
                        Advanced Memory Retrieval Techniques \cite{zhou2023agents,feng2023llama,nottingham2023selective,piterbarg2024diff,wu2023deciphering},\\
                        Video-Language Model for Task-Relevant Encoding \cite{fan2022minedojo}
                        }, 
					    text width=\textwidthexample, 
					    node options={align=left}, 
					    rounded corners=false,
					    fill=\fillexample]
                    ]
				]
                [\textbf{Learning} \S\ref{subsec:learning}, text width=\textwidthsubsection, for tree={fill=\fillsubsubsection}, minimum height=\miniheightsubsection, fill=\fillsubsection
                    [{
                    Iterative and Feedback-Based Learning~\cite{wang2023voyager,qian2023communicative,schaferVisualEncodersDataEfficient2023},\\
                    Reinforcement and Offline Learning~\cite{yang2023octopus,zhang2023sprint,lan2023llm},\\
                    Experience-Based and Memory Utilization Learning~\cite{xu2023exploring,ma2023large},\\
                    Collaborative and Adaptive Learning~\cite{dingDesignGPTMultiAgentCollaboration2023,chen2023autoagents},\\
                    Simulated and Contextual Learning~\cite{park2022social,wu2023tidybot}},
					text width=\textwidthsubsubsection,
                    rounded corners=false,
					node options={align=left}
					]
                ]
                [\textbf{Reasoning} \S\ref{subsec:reason}, text width=\textwidthsubsection, for tree={fill=\fillsubsubsection}, minimum height=\miniheightsubsection, fill=\fillsubsection
                    [
					{Strategic Reasoning \cite{park2023generative,wangAvalonGameThoughts2023}, Task-oriented Reasoning \cite{zhang2023sprint,anonymous2023ghost},\\
                    Contextual and Adaptive Reasoning \cite{wang2023voyager,huangInner},\\
                    Multi-agent Collaboration and Interactive Reasoning \cite{hong2023metagpt,qian2023communicative}},
					text width=\textwidthsubsubsection,
                    rounded corners=false,
					node options={align=left}
					]
                ]
                [\textbf{Decision-making} \S\ref{subsec:decision}, text width=\textwidthsubsection, for tree={fill=\fillsubsubsection}, minimum height=\miniheightsubsection, fill=\fillsubsection
                    [
					\textbf{Multi-Hop Inference},
					text width=\textwidthsubsubsectionrealtitle,
					node options={align=center}
                        [{Collaborative Multi-Agent Systems \cite{hong2023metagpt,qian2023communicative,chen2023agentverse},\\
                        Multi-Level Decision-Making \cite{yuan2023skill,zhang2023proagent,cai2023groot},\\
                        Structured Process and Management \cite{zhou2023agents,wang2023describe}}, 
					    text width=\textwidthexample, 
					    node options={align=left}, 
					    rounded corners=false,
					    fill=\fillexample]
					]
                    [
                    \textbf{Long-term Planning},
					text width=\textwidthsubsubsectionrealtitle,
					node options={align=center}
					    [{CoT Planning~\cite{wu2023visual}, Skill Pre-training and Adaptive Learning~\cite{zhang2023sprint,zhaiBuildingOpenEndedEmbodied2023},\\
                        Task Decomposition into Manageable Subtasks\cite{dingDesignGPTMultiAgentCollaboration2023,jin2023alphablock},\\
                        Structured Planning and Iterative Processes~\cite{hong2023metagpt,qian2023communicative,chen2023agentverse}}, 
					    text width=\textwidthexample, 
					    node options={align=left}, 
					    rounded corners=false,
					    fill=\fillexample]
                    ]
                ]
                [\textbf{Reflection} \S\ref{subsec:reflect}, text width=\textwidthsubsection, for tree={fill=\fillsubsubsection}, minimum height=\miniheightsubsection, fill=\fillsubsection
                    [{Feedback-Based Self-Improvement~\cite{wang2023voyager,anonymous2023ghost,chen2023agentverse,hong2023metagpt,lan2023llm},\\
                    Iterative Optimization of Plans and Actions~\cite{huangInner,zhaiBuildingOpenEndedEmbodied2023,cai2023groot},\\
                    Collaboration and Multi-Agent Interaction~\cite{qian2023communicative,wangAvalonGameThoughts2023},\\
                    Debate and Theory of Mind-Based Reflection~\cite{duImprovingFactualityReasoning2023,guo2023suspicion} },
					text width=\textwidthsubsubsection,
                    rounded corners=false,
					node options={align=left}
					]
                ]
			]
            [\textbf{Action} \S\ref{sec:action}, text width=\textwidthsection, minimum height=\miniheightsection, for tree={fill=\fillsection}
                [\textbf{Behavior} \S\ref{subsec:behave}, text width=\textwidthsubsection, for tree={fill=\fillsubsubsection}, minimum height=\miniheightsubsection, fill=\fillsubsection
                    [{Iterative Prompting with Environmental Feedback \cite{wang2023voyager},\\
                    Role-Specific Prompt for Program Generation \cite{zhang2023sprint,dingDesignGPTMultiAgentCollaboration2023},\\
                    Natural Language Instruction Interpretation \cite{rana2023sayplan,wu2023embodied}},
					text width=\textwidthsubsubsection,
                    rounded corners=false,
					node options={align=left}
					]
                ]
                [\textbf{Dialogue} \S\ref{subsec:dialogue}, text width=\textwidthsubsection, for tree={fill=\fillsubsubsection}, minimum height=\miniheightsubsection, fill=\fillsubsection
                    [{\textbf{Agent-agent Interactions:},\\
                    Collaborative Frameworks~\cite{chen2023autoagents},
                    Conversation-driven Control~\cite{wu2023autogen, zhou2023agents, fuImprovingLanguageModel2023},\\
                    Game-based Strategic Behaviors~\cite{xu2023exploring, light2023from,abdelnabi2023llmdeliberation}},
					text width=\textwidthsubsubsection,
                    rounded corners=false,
					node options={align=left}
					]
                    [
                    {\textbf{Human-agent Interactions:},\\
                    Instruction-following and Collaborative Task Solving~\cite{li2023camel, park2023generative, qian2023communicative},\\
                    Gameplay Enhancement and Creative Assistance~\cite{akoury2023framework, zhu2023calypso, dingDesignGPTMultiAgentCollaboration2023},\\
                    Natural and Flexible Communication~\cite{yang2023octopus}},
					text width=\textwidthsubsubsection,
                    rounded corners=false,
					node options={align=left}
                    ]
                ]
                [\textbf{Consistency} \S\ref{subsec:consistent}, text width=\textwidthsubsection, for tree={fill=\fillsubsubsection}, minimum height=\miniheightsubsection, fill=\fillsubsection
                    [{Specific Strategies~\cite{gong2023mindagent,wu2023visual,xuLanguageAgentsReinforcement2023,wang2023selfconsistency,hu2024pokellmon},\\
                    Structural Approaches~\cite{wu2023spring,wang2023voyager,anonymous2023ghost,hong2023metagpt,zhou2023agents},\\
                    Feedback and Adaptation Mechanisms~\cite{feng2023llama,wu2023autogen},
                    Neuro-Symbolic Approach \cite{dong2023cottage}},
					text width=\textwidthsubsubsection,
                    rounded corners=false,
					node options={align=left}
					]
                ]
            ]
            [\textbf{Challenges Throughout the Playing Cycle} \S\ref{sec:other_challenges}, text width=\textwidthsection, minimum height=\miniheightsection, for tree={fill=\fillsection}
                [\textbf{Hallucination} \S\ref{subsec:hallu}, text width=\textwidthsubsection, for tree={fill=\fillsubsubsection}, minimum height=\miniheightsubsection, fill=\fillsubsection
                    [
					{Structured Reasoning and Contextual Awareness~\cite{hong2023metagpt,wang2023jarvis,wu2023spring,qian2023communicative,xu2023exploring,chen2023gamegpt},\\ 
                    Interactive and Collaborative Methods \cite{zhu2023calypso,chen2023agentverse},\\
                    Leveraging External Knowledge and Advanced Learning Techniques~\cite{hu2024pokellmon,wu2023visual,wangApolloOracleRetrievalAugmented2023},\\
                    Specific Prompt or Feedback Mechanisms~\cite{HumanlevelPlayGame2022,dingDesignGPTMultiAgentCollaboration2023,wangRoleLLMBenchmarkingEliciting2023,gong2023mindagent},\\
                    Refinement and Fine-Tuning Methods~\cite{duImprovingFactualityReasoning2023,jin2023alphablock,wu2023embodied}},
					text width=\textwidthsubsubsection,
                    rounded corners=false,
					node options={align=left}
					]
                ]
                [\textbf{Error Correction} \S\ref{subsec:error}, text width=\textwidthsubsection, for tree={fill=\fillsubsubsection}, minimum height=\miniheightsubsection, fill=\fillsubsection
                    [
					{Iterative Learning and Feedback Mechanisms~\cite{li2023camel,wang2023voyager,hong2023metagpt},\\
                    Real-time Feedback and Learning from Mistakes~\cite{rana2023sayplan, wang2023jarvis,lan2023llm},\\
                    Action Validation and Error Messaging \cite{gong2023mindagent},\\
                    Accuracy, Consistency, and Relevance in Feedback \cite{li2023auto},\\
                    Peer Review and System Testing Approaches~\cite{qian2023communicative,chen2023agentverse},\\
                    Structured Systems for Error Minimization~\cite{xu2023exploring,wangApolloOracleRetrievalAugmented2023,duImprovingFactualityReasoning2023},\\
                    Strategy Implementations for Error Correction~\cite{yang2023octopus,jin2023alphablock}},
					text width=\textwidthsubsubsection,
                    rounded corners=false,
					node options={align=left}
					]
                ]
                [\textbf{Generalization} \S\ref{subsec:gene}, text width=\textwidthsubsection, for tree={fill=\fillsubsubsection}, minimum height=\miniheightsubsection, fill=\fillsubsection
                    [
					{Context-Guided Reasoning and Long-Term Planning~\cite{wu2023spring,anonymous2023ghost,zhang2023building},\\
                    Zero-Shot Generalization and Role Adaptability  \cite{wang2023voyager,wang2023describe,wangRoleLLMBenchmarkingEliciting2023},\\
                    Adaptive Frameworks for Software Development and Task Planning~\cite{qian2023communicative,jin2023alphablock},\\
                    Pre-training and Skill Transfer for New Environments~\cite{zhang2023sprint},\\
                    Scalable Human-Like Interface~\cite{tan2024towards,sima}},
					text width=\textwidthsubsubsection,
                    rounded corners=false,
					node options={align=left}
					]
                ]
                [\textbf{Interpretability} \S\ref{subsec:inter}, text width=\textwidthsubsection, for tree={fill=\fillsubsubsection}, minimum height=\miniheightsubsection, fill=\fillsubsection
                    [
					{COT Reasoning and Decision-Making Transparency~\cite{wu2023spring,feng2023llama,park2023generative,jin2023alphablock},\\
                    Structured and Modular Decision-Making Approaches~\cite{wang2023voyager,anonymous2023ghost,zhang2023building},\\
                    Communication and Role-Based Interpretability \cite{hong2023metagpt,chen2023agentverse,qian2023communicative,wu2023autogen,wangAvalonGameThoughts2023},\\
                    Techniques for Detailed Action Explanation~\cite{wang2023describe,schaferVisualEncodersDataEfficient2023,wu2023visual},\\
                    Explicit Role and Action Definition for Clarity~\cite{wangApolloOracleRetrievalAugmented2023,xuLanguageAgentsReinforcement2023,huang2023ark,cai2023groot}},
					text width=\textwidthsubsubsection,
                    rounded corners=false,
					node options={align=left}
					]
                ]
            ]
            [\textbf{Conclusion and Future Direction} \S\ref{sec:Future Direction}, text width=\textwidthsection, minimum height=\miniheightsection, for tree={fill=\fillsection}]
		]
	\end{forest}
	\caption{Taxonomy of research on complex digital game playing agents and large models.}
    \label{fig:taxonomy}
    \vspace{-0.3cm}
\end{figure*}